\documentclass[sigconf]{acmart}

\usepackage{algorithm}
\usepackage{algorithmic}
\usepackage{hyperref}
\usepackage{xcolor}
\usepackage{enumitem}
\usepackage{amsmath,amsfonts,bm}
\usepackage{multirow}
\usepackage{makecell}
\usepackage{tabularx}
\usepackage{lipsum}
\usepackage{array}
\usepackage{xcolor}

\usepackage{CJK}  
\usepackage{xcolor}  

\AtBeginDocument{%
  }

\setcopyright{acmlicensed}
\copyrightyear{2018}
\acmYear{2018}
\acmDOI{XXXXXXX.XXXXXXX}
\acmConference[Conference acronym 'XX]{Make sure to enter the correct
  conference title from your rights confirmation emai}{June 03--05,
  2018}{Woodstock, NY}
\acmISBN{978-1-4503-XXXX-X/18/06}

\newsavebox{\tablebox}
\begin{document}

\title{Constructing a Question-Answering Simulator through the Distillation of LLMs}

\author{Haipeng Liu}
\email{liuhp22@mails.jlu.edu.cn}
\affiliation{%
  \institution{School of Artificial Intelligence, Jilin University}
 \country{China}
}

\author{Ting Long}
\email{longting@jlu.edu.cn}
\affiliation{%
  \institution{School of Artificial Intelligence, Jilin University}
  \country{China}
}

\author{Jing Fu}
\email{fulei24@mails.jlu.edu.cn}
\affiliation{%
  \institution{School of Artificial Intelligence, Jilin University}
  \country{China}
}








\renewcommand\footnotetextcopyrightpermission[1]{}

\newcommand{\ltc}[1]{\begin{CJK}{UTF8}{gbsn}\textcolor{red}{修改意见：#1}\end{CJK}}
\newcommand{\lts}[1]{\begin{CJK}{UTF8}{gbsn}\textcolor{brown}{修改意见：#1}\end{CJK}}
\newcommand{\todo}[1]{\begin{CJK}{UTF8}{gbsn}\textcolor{pink}{todo：#1}\end{CJK}}
\newcommand{\lhp}[1]{\begin{CJK}{UTF8}{gbsn}\textcolor{orange}{修改：#1}\end{CJK}}

\newcommand{\ie}{\textit{i}.\textit{e}.}
\newcommand{\eg}{\textit{e}.\textit{g}.}

\settopmatter{printacmref=false} 

\begin{abstract}
  The question-answering (QA) simulator is a model that mimics real student learning behaviors and predicts their correctness of their responses to questions. QA simulators enable educational recommender systems (ERS) to collect large amounts of training data without interacting with real students, thereby preventing harmful recommendations made by an undertrained ERS from undermining actual student learning. Given the QA history, there are two categories of solutions to predict the correctness, conducting the simulation: (1) LLM-free methods, which apply a traditional sequential model to transfer the QA history into a vector representation first, and make predictions based on the representation; (2) LLM-based methods, which leverage the domain knowledge and reasoning capability of LLM to enhence the prediction. LLM-free methods offer fast inference but generally yield suboptimal performance. In contrast, most LLM-based methods achieve better results, but at the cost of slower inference speed and higher GPU memory consumption. In this paper, we propose a method named LLM Distillation
based Simulator (LDSim), which distills domain knowledge and reasoning capability from an LLM to better assist prediction, thereby improving simulation performance. Extensive experiments demonstrate that our LDSim achieves strong results on both the simulation task and the knowledge tracing (KT) task. Our code is publicly available at \url{https://anonymous.4open.science/r/LDSim-05A9}.

\end{abstract}



\received{20 February 2007}
\received[revised]{12 March 2009}
\received[accepted]{5 June 2009}

\maketitle

\vspace{-10pt}
\section{Introduction}
\begin{figure}[t]
\centering
\includegraphics[width=1\columnwidth]{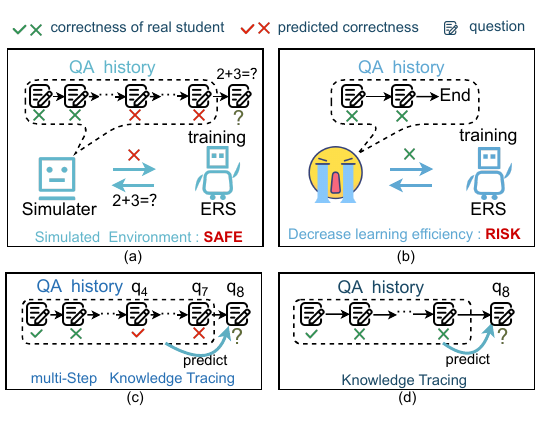}
\vspace{-15pt}
\caption{Illustration of the role and tasks of the QA simulator.} 

\label{fig: intro}
\end{figure}

A Question-Answering (QA) Simulator \cite{long2025simulating, gao2025agent4edu} is a model that captures the learning behavior of students from their question-answering (QA) history, and simulates the response of students to interact with an educational recommender system (ERS) \cite{chen2023set, liu2024hierllm, kubotani2021rltutor}, ensuring the training safety of ERS. As it is illustrated in Figure \ref{fig: intro}(a) and (b), the training of ERS requires a large number of samples that include the recommendation of ERS and responses of students. As a result, the ERS has to interact with students to collect enough training data and conduct trial-and-error to identify the efficient recommendation strategy. However, the unconverged ERS carries the risk of generating random recommendations that diminish students' learning efficiency and experience, which goes against the original intention of using ERS in online education to enhance both learning efficiency and learning experience. The QA simulator acts as a simulated student to interact with the ERS, enabling the ERS to collect training data and train in a simulated environment, thereby mitigating the potential risk of degrading students' learning efficiency and experience.

Most existing simulators are constructed based on multi-step knowledge tracing (KT). Given the QA history of a student and a sequence of the subsequent questions, the simulator simulates students' QA behavior by consecutively predicting the correctness of responses to those questions, where predictions are conditioned on the outcomes of its previous predictions. As it is illustrated in Figure \ref{fig: intro} (d), 
when simulating a student's correctness on question $q_8$, the simulator cannot rely on the human student's actual QA history from step $1$ to $7$ to make prediction, as the actual correctness of response from $q_4$ to $q_7$ is unknown during interaction with the ERS. Instead, the simulator make prediction based on the partially synthetic QA history to make prediction, in which the correctness from $q_4$ to $q_7$ is replaced by the predicted correctness.

Under the above setting, the solutions to simulator can be broadly categorized into LLM-free methods and LLM-based methods.
LLM-free methods typically employ traditional sequential models, such as recurrent neural networks \cite{hochreiter1997long} and Transformers \cite{vaswani2017attention}, to encode the QA history into a hidden state and perform multi-step predictions based on the state. These methods are lightweight, require less GPU memory, and offer faster inference speed. However, their performance is often suboptimal. When used for ERS interaction, LLM-free simulators may mislead the ERS, resulting in recommendations that perform well for the simulator but provide little benefit to real students. 
LLM-based methods, on the other hand, feed the QA history and related information into a LLM in the form of prompts, followed by multi-step prediction through fine-tuning \cite{wang2025llm} or prompt engineering \cite{gao2025agent4edu}. Benefiting from the strong domain knowledge and reasoning capabilities of LLMs, such methods can better capture complex patterns in QA history, thereby achieving more accurate predictions. Nonetheless, LLM-based methods are computationally expensive: they involve large parameter counts, demand high GPU memory, and have slower inference speeds, making them inefficient and costly for use in real-time ERS interaction to support ERS training.

To address the issue, we propose a method called LLM Distillation based Simulator (LDSim), which distills the domain knowledge and reasoning capabilities of LLMs into a lightweight network to conduct the multi-step prediction of the simulator. Specifically, LDSim consists of three modules: knowledge distillation module (KD), reasoning distillation module (RD) and Simulation module (Sim). The KD distills knowledge about the prerequisite relation of concepts from LLM. The RD distills the reasoning of students' mastery of concepts at each time step. The Sim devices the neural network to leverage the distilled information to conduct the multi-step prediction and simulate students' QA behavior. To validate the performance of LDSim, we conduct extensive experiments and compare it with 10 outstanding baselines. The experiment results demonstrate that our LDSim is effective and efficient in simulation. 

In summary, the contributions of this work are:
\begin{itemize}[leftmargin=15pt]
    \item We propose LDSim, a method that distills LLMs for the construction of QA simulators. To the best of our knowledge, this is the first attempt to apply LLM distillation in the educational domain.
    \item Our LDSim integrates the strengths of both LLM-based and LLM-free methods, achieving high performance in simulation tasks while remaining memory-efficient and computationally efficient.
    \item Extensive experiments demonstrate that our LDSim  outperforms outstanding baselines in the simulation task. 
\end{itemize}

\section{Related Work}
Most existing QA simulators are constructed based on knowledge tracing (KT) models. According to their methods of encoding questions and concepts, we categorize them into two types: LLM-free methods and LLM-based methods.

\subsection{LLM-free methods}
LLM-free simulators are built based on traditional sequential models, such as RNN and Transformer. They first feed the ID of questions, concepts, and the correctness of students' responses in QA history to RNN (Transformer) to model students' learning states, and then use the learning states to make predictions. For example, Deep Knowledge Tracing (DKT) \cite{piech2015deep} is one of the most representative KT methods that can be used as a simulator. It applies an LSTM \cite{hochreiter1997long} to encode the sequence of QA history, in which the correctness labels are replaced with the previously predicted results, and predicts the correctness of the next response based on the LSTM's hidden state. Considering that questions of varying difficulty induce different cognitive biases which affect performance, DisKT \cite{zhou2025disentangled} introduces a mechanism to model easy and hard questions separately. To enhance the robustness and generalization, ATKT \cite{aktguo2021enhancing} introduces adversarial training, while DSim \cite{long2025simulating} introduces the conditional diffusion models.
Although LLM-free simulators have been widely adopted and improved upon, they suffer from an inherent limitation: they treat questions and concepts as discrete, independent IDs, which overlooks the semantic and relational information among them, limiting the performance of simulation.

\subsection{LLM-based methods}
LLM based methods are built based on LLM, which leverages the open-world knowledge and reasoning capabilities to enhance the prediction. For example, SINKT \cite{fu2024sinkt} utilizes the domain knowledge of LLMs to construct a hierachical graph to encode the questions and concepts, thus enhancing the prediction. Similarly, LLM-KT \cite{wang2025llm} leveraging the open-world knowledge of LLM to encoding the concepts for better prediction. Agent4Edu \cite{gao2025agent4edu} takes advantage of LLMs' reasoning ability by allowing the LLM to directly predict student responses.
Compared to LLM-free methods, LLM-based methods can better capture complex pattern in QA history and semantic information among concepts (questions), enabling more outstanding performance. 
However, their large parameter size leads to significant computational and runtime costs, which severely limit their applicability in simulation as well as real-world scenarios.

To maintain the performance of LLM-based methods and the efficiency of LLM-free methods, we propose LDSim, which distills the domain knowledge and reasoning capabilities of LLMs into a lightweight neural network, ensuring both high simulation accuracy and computational efficiency.

\section{Problem Definition}

In an online learning site, let $\mathcal{U}$ denote the set of students, $\mathcal{Q}$ the set of questions, and $\mathcal{C}$ the set of concepts. For an arbitrary student $u \in \mathcal{U}$ learning in the site, we first define their QA record as follows:
\begin{definition}
    (\textbf{QA record}). The QA record of a student $u \in \mathcal{U}$ at time step $i$ consists of the question attempted by the student at step $i$, the set of concepts associated with the question, and the student's response.
\end{definition}
\begin{equation}
    x_i^u = (q_i, \mathcal{C}_i, r_i),
\end{equation}
where $q_i \in \mathcal{Q}$ is the question attempted at time step $i$, $\mathcal{C}_i \subseteq \mathcal{C}$, with $\mathcal{C}_i \neq \emptyset$, represents the set of concepts involved in question $q_i$, and $r_i \in \{0, 1\}$ indicates whether the student answered the question correctly (with $r_i = 1$ for a correct response, and $r_i = 0$ for an incorrect one).
For example, if student $u$ attempted the question "2+6-4 = ?" at step 5, the corresponding concepts are $\mathcal{C}_5 = \{\text{"addition"}, \text{"subtraction"}\}$. If the student answers this problem correctly, then $r_5 = 1$.

Then we define the student $u$'s QA history as:
\begin{definition}
(\textbf{Question-answering (QA) history}). The QA history of a student $u \in \mathcal{U}$ at time step $t$ refers to the QA records prior to time step $t$.

\begin{equation}
    H_t^u = \{(x_1^u, x_2^u, \ldots, x_t^u,\},
\end{equation}
\end{definition}

Given a student $u$'s QA history $H_t^u$ at step $t$, this work aims to simulate the correctness of $u$'s responses to the subsequent questions recommended by ERS in the subsequent steps.
That is, we aim to build a QA simulator that, when the ERS recommends a sequence of questions $q_{t+1}, q_{t+2}, ..., q_{t+n}$ to the student, can predicts the student's responses $\hat{r}_{t+1}, \hat{r}_{t+2}, ..., \hat{r}_{t+n}$ based on the student's QA history $H_t^u$, the recommended questions $q_{t+1}, q_{t+2}, ..., q_{t+n}$, and the corresponding concepts $\mathcal{C}_{t+1}, \mathcal{C}_{t+2}, ..., \mathcal{C}_{t+n}$:
\begin{equation}
\label{prob_def}
    \hat{r}_{t+i}|_{i=1}^{n} = f_s(H_t^u, q_{t+i}|_{i=1}^{n}, \mathcal{C}_{t+i}|_{i=1}^{n}),
\end{equation}
where $f_s$ is the QA simulator.

\section{Methodology}
To build a simulator to simulate students' QA
behavior and interact with ERS, we propose a method called LLM Distillation based Simulator (LDSim).

\begin{figure*}[t]
\centering

\includegraphics[width=1.95\columnwidth]{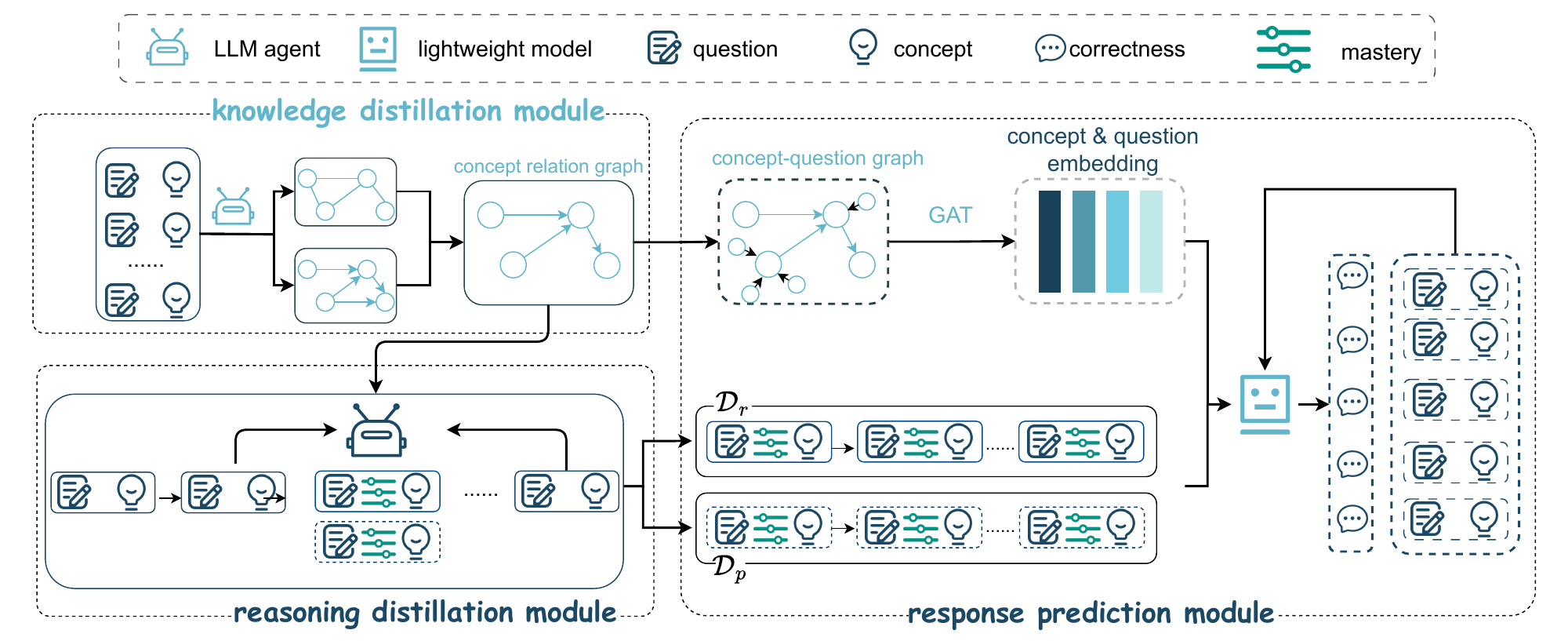}
\caption{The pipeline of our method.}
\label{framework}
\end{figure*}

As illustrated in the Figure \ref{framework}, 
LDSim consists of three modules: the Knowledge Distillation Module (KD), the Reasoning Distillation Module(RD), and the Simulation Module (SiM). Both KD and RD are built based on LLM, which distill the domain knowledge and reasoning capability of LLM to formulate the concept relation graph and the distilled data, respectively. The Sim is a lightweight neural network which learn with the graph and the distilled data to replicate the performance of LLM-based methods in simulation, maintains the efficiency of LLM-free methods at the same time. In the following, we will discuss the KD and RD first, and subsequently the Sim.

\subsection{Knowledge Distillation Module}
\label{sec:KD}
The world knowledge distillation module is responsible for transforming the LLM's world knowledge about prerequisite relationships among concepts into a concept relation graph. For example, in elementary mathematics, students must have a solid mastery of addition before learning multiplication to make the learning process easier. Therefore, addition is a prerequisite of multiplication. To distill the concept relation graph, we design a two-stage LLM-based inference strategy. Specifically, we adopt the following steps:

\begin{figure}
    \centering
    \includegraphics[width=1.0\linewidth]{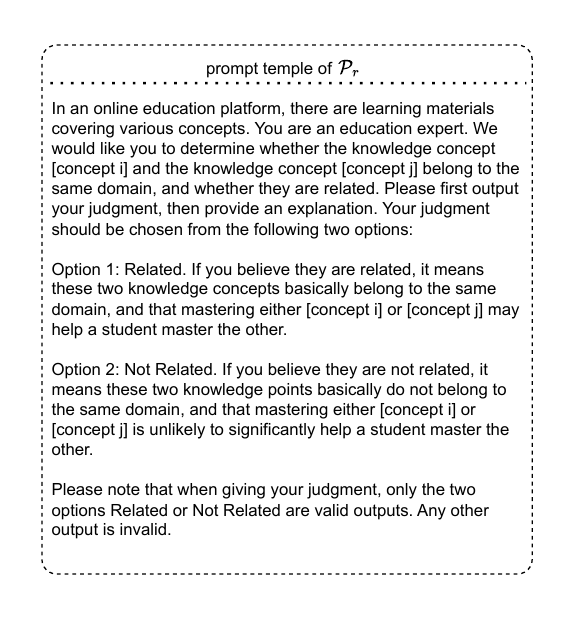}
    \caption{prompt template of $\mathcal{P}_r$, where [concept i] and [concept j] correspond to $\text{TEXT}(c_i)$ and $\text{TEXT}(c_j)$ in Equation \ref{eq:relevance}.}

    \label{fig:rela_prompt}
\end{figure}

First, given two arbitrary concepts $c_i$ and $c_j$,  we instruct an LLM to assess whether the given pair of concepts is related. We design the prompt template as it is illustrated in Figure \ref{fig:rela_prompt}, encapsulating the concepts into the prompt template and feeding to LLM to obtain the assessment result:
\begin{equation} \label{eq:relevance}
    b_{i,j}^r = \text{LLM}_r \left (\mathcal{P}_r( \text{TEXT}(c_i), \text{TEXT}(c_j)\right),
\end{equation}
where $\mathcal{P}_r$ is the prompt template used for assessing the relevance of two arbitrary concepts, and $\text{TEXT}(c_i), \text{TEXT}(c_j)$ are the textual descriptions of concept $c_i$ and $c_j$, respectively. $b_{i,j}^r \in \{0, 1\}$ indicates whether concept $c_i$ and $c_j$ is relevant, where $b_{i,j}^r = 1$ means they are relevant, and $b_{i,j}^r = 0$ otherwise. $\text{LLM}_r$ denotes the LLM.
To reduce the inherent randomness in LLM generation \cite{atil2024llm, ouyang2025empirical} and improve the accuracy of relevance assessment, we swap the position of $c_i$ and $c_j$ in prompt template, and conduct Eq.~\ref{eq:relevance} again, and obtain $b_{j,i}^r$ respectively.

\begin{figure}
    \centering
    \includegraphics[width=1.0\linewidth]{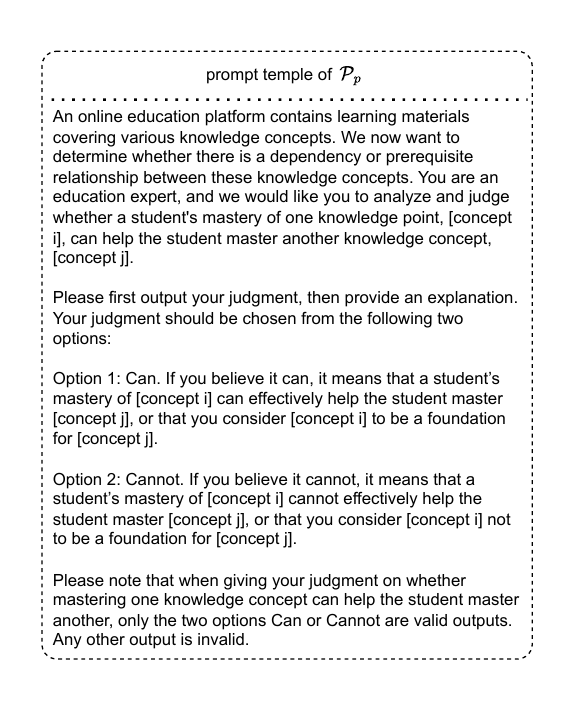}
    \caption{prompt template of $\mathcal{P}_p$, where [concept i] and [concept j] correspond to $\text{TEXT}(c_i)$ and $\text{TEXT}(c_j)$ in Equation \ref{eq:requisite}. }
    \label{fig:pre_prompt}
\end{figure}

Next, we instruct an LLM to assess the requisite relation if both $b_{i,j}^r = 1$ and $b_{j, i}^r = 1$. We design a prompt template $\mathcal{P}_p$ for requisite relation identification, which is illustrated in Figure \ref{fig:pre_prompt}. Then, we encapsulate the concept $c_i$ and $c_j$ into the prompt template, feed the prompt to LLM:
\begin{equation} \label{eq:requisite}
    b_{i,j}^p = \text{LLM}_p \left(\mathcal{P}_p(\text{TEXT}(c_i), \text{TEXT}(c_j)\right),
\end{equation}
where $\mathcal{P}_p$ is the template prompt used to determine the prerequisite relation between concept $c_i$ and $c_j$. 
$b_{i,j}^p \in \{0, 1\}$ indicates whether $c_j$ is a prerequisite concept of $c_i$, where $b_{i,j}^p = 1$ means $c_j$ is a prerequisite concept of $c_i$, and $b_{i,j}^p = 0$ otherwise. $\text{LLM}_r$ denotes the LLM. Since $b_{i,j}^p = 0$ only indicates that $c_j$ is not a prerequisite of $c_i$, it is still possible that $c_i$ is a prerequisite of $c_j$. Therefore, we swap the positions of $c_i$ and $c_j$ in the prompt template and conduct the assessment in Eq.~\ref{eq:requisite} again.

Finally, we construct the concept relation graph $\mathcal{G}_c = (\mathcal{V}_c, \mathcal{E}_c)$ according to the prerequisite relation obtained by Eq.~\ref{eq:requisite}. Here, $\mathcal{V}_c$ denotes the set of nodes in the concept relation graph $\mathcal{G}_c$, and each node represents a concept. $\mathcal{E}_c$ is the set of edges in $\mathcal{G}_c$. For arbitrary concepts $c_i$ and $c_j$, if $b_{i,j}^p = 1$, there is a edge $\langle c_i, c_j \rangle \in \mathcal{E}_c$, where the start node is $c_i$, and the end node is $c_j$.

\subsection{Reasoning Distillation Module}
Researches has shown that a student's latent mastery of the concepts involved in a question is a key factor influencing whether the student can correctly answer the question \cite{lord2012applications, aktguo2021enhancing}. Therefore, we design the reasoning distillation module to distill the LLM's reasoning ability in inferring students' mastery on concepts into training data, supporting the further training of a lightweight model (discussed in section \ref{sec:sim}). Specifically, we take the following steps to obtain the distilled training data:

\begin{figure}
    \centering
    \includegraphics[width=1.0\linewidth]{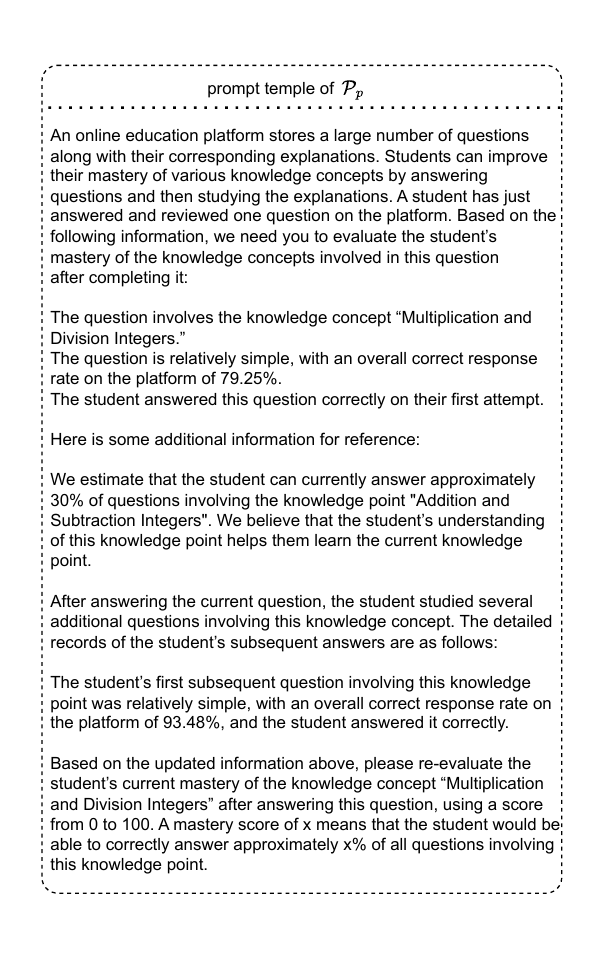}
    \caption{An example of $\mathcal{P}_m$, where the mastery information in the prompt at step $i$ comes from $m_t|_{t=1}^{i-1}$. 
}

    \label{fig:mas_prompt}
\end{figure}

Firstly, for any QA record $(q_i, \mathcal{C}_i, r_i)$ of a student, the prompt template $\mathcal{P}_m$ encapsulates all the student's QA records before and after step $i$, along with the correct rate that questions and concept relation graph $\mathcal{G}$ (Section \ref{sec:KD}). It is used to guide the LLM in reasoning about the student's mastery of the concepts in $\mathcal{C}_i$.
\begin{equation} \label{eq:distill_mastery}
	m_i, s_i = LLM_g \left(\mathcal{P}_m \left( \mathcal{G}, (q_t, \mathcal{C}_t, r_t)|_{t=1}^T \right) \right),
\end{equation}
where $\mathcal{P}_m$ is the prompt template used for distilling the mastery (an example is shown in Figure \ref{fig:mas_prompt}). $m_t \in [0, 1]$ represents the average mastery of student $u$ over all concepts in $\mathcal{C}_i$ when answering question $q_i$ at step $i$, with a higher $m_i$ indicating a better overall mastery of $\mathcal{C}_i$ at that time. $\text{LLM}_g$ denotes the LLM. $s_i$ is the credit score of LLM.

For the QA records of student $u$, we conduct the operation in Eq.~\ref{eq:distill_mastery}. Next, we integrate the mastery infered by LLM to the corresponding QA records, and formulate the distilled QA records of student $u$:  $(q_t, \mathcal{C}_t, r_t, m_t, s_t) |_{t=1}^T$. Conducting the same operation on the QA records of all the students in the training data, we obtain the distilled the data of training dataset. We denote the data as actual distilled trainning dataset $\mathcal{D}_r$.

The dataset $\mathcal{D}_r$ obtained above only distill from students actual QA records, where the mastery of concepts covered by the actually answered question. However, as each students only answer one question at a specific time, dataset $\mathcal{D}_r$ contains limited information about LLM reasoning capability in inferring students' mastery of concepts.  To fullly distill the power of LLM reasoning capabiliy, we conduct an augmention strategy. Specifically, for QA record $(q_t, \mathcal{C}_t, r_t)$ of student $u$, we randomly select $n$ questions from  $\mathcal{Q}$ and synthesize pseudo QA (PQA) records $(q^i_t, \mathcal{C}^i_t, r^i_t, m^i_t, s^i_t) |_{i=1}^n$ by conducting the operation in Eq.~\ref{eq:distill_mastery}, assuming the student answered the another $n$ questions in parallelized step. For each PQA record $(q^i_t, \mathcal{C}^i_t, r^i_t, m^i_t, s_t^i) $,  $q^i_t$ is the randomly selected question. $\mathcal{C}^i_t$ is the concepts covered by question $q^i_t$. $r^i_t$ is set to $-1$, as the correctness of student correctly answer  $q^i_t$  is unknown. $m^i_t$ is the average mastery of concepts in  $\mathcal{C}^i_t$ inferred by LLM. $s^i_t$ is the credict score of LLM. We denote the dataset distilled from the pseudo QA records as $\mathcal{D}_p$.

\subsection{Simulation Module}
\label{sec:sim}
The simulation module is a light-weight neural network, which is responsible to learn from the distilled concept relation graph and training data $\mathcal{D}$, simulating students' QA behavior. In the following, we will discuss how the simulation module makes predictions to simulate students' QA behavior first, then we will discuss how to optimize the simulation module.

\paragraph{\textbf{Correctness prediction}} Given an arbitrary question $q_t$ and the QA history of a student at step $t$, we encode the question, concept and QA history first. Then, we estimate the student's mastery level of question-covered concepts $C_t$ based on the encoding result to predict the correctness of the response. Specifically, we adopt the following steps:

To encode the questions and concepts, we construct a hierarchical concept-question graph $\mathcal{G}$, $\mathcal{G} = \{\mathcal{V}, \mathcal{E}\}$. Here, $\mathcal{V} = \mathcal{C} \cup \mathcal{Q}$, representing the set of concepts and questions. $\mathcal{E}$ represents the edges, representing the relations among nodes. There are two types of edges in $\mathcal{E}$: (1) $\mathcal{E}_c$, which indicates the prerequisite relations among concepts (obtained by knowledge distillation module in section \ref{sec:KD}); (2) the relations between concepts and questions: If concept $c_i$ is covered by question $q_j$, there is an edge between concept $c_i$ and $q_j$.

Next, we apply Graph Attention Network (GAT) \cite{velivckovic2017graph} on the hierarchical concept-question graph $\mathcal{G}$ to encode the semantic embedding of concepts and questions:
\begin{equation}
\label{GAT}
    \bm{E}^q, \bm{E}^c = \text{GAT}(\mathcal{G}), 
\end{equation}
where $\bm{E}^q \in \mathbb{R}^{|\mathcal{Q}| \times d}$, $\bm{E}^c \in \mathbb{R}^{|\mathcal{C}| \times d}$. $\bm{e}^q_i \in \bm{E}^q$ is a row vector of $\bm{E}^q$, representing the semantic embedding of concept $q_i$, $\bm{e}^c_i \in \bm{E}^c$ is a row vector of $\bm{E}^c$, representing the semantic embedding of concept $c_i$.

To encode the QA history, we first encode any arbitrary concept $c_j$ covered by the question in the QA history:

\begin{equation}
\label{DR1}
\begin{aligned}
    & \bm{u}_j = f_1(\text{cor}(c_j)) \oplus f_2(\text{cou}(c_j)) \\
    & \bm{z}_j = Attn(\bm{e}^c_j, \bm{u}_j, \bm{u}_j)
\end{aligned}
\end{equation}
where $\text{cor}(c_j)$ and $\text{cou}(c_j)$ denote the student's historical correctness rate and the number of attempts for concept $c_j$, respectively, based on the QA records in the training data. The functions $f_1(\cdot)$ and $f_2(\cdot)$ are linear layers used to project scalar inputs into vectors. The symbol $\oplus$ denotes vector concatenation. $\bm{e}^c_j$ is the semantic embedding of concept $c_j$ obtained from Eq.~\ref{GAT}. $Attn(\cdot, \cdot, \cdot)$ represents multi-head attention, where the query is $\bm{e}^c_j$, and the key and value are $\bm{u}_j$. Since $\bm{z}_j$ contains the information of the historical correct rate and trail, we denote $\bm{z}_j$ as the contextual embedding of concept $c_j$.


With Equation \ref{DR1}, we can represent the concept set $C_i$ at any arbitrary step $i$ with the mean contextual embedding of all concepts within the set:

\begin{equation}
\label{DR2}
    \bm{\bar{z}}_t = mean(\bm{z}_t|c_t \in \mathcal{C}_t).
\end{equation}

Subsequently, we leverage the questions and corresponding correctness in the QA history to encode the QA history as the learning state at step $t$:
\begin{equation}
\label{learning state}
\begin{aligned}
    & \bm{u_i^q} = \bm{e_i^q}+f_3(\text{cor}(q_i)) \\
    & \bm{s}_t = Attn(\bm{\bar{z}}_t, \bm{e_i^q}|_{i=1}^{t-1}, \bm{u_i^q}|_{i=1}^{t-1}),
\end{aligned}
\end{equation}
where $\text{cor}(q_i)$ denotes the correct rate that question $q_t$ is correctly answered in the ITS
, which is estimated based on the training data. $\bm{e_t^q}$ is the semantic embedding of question $q_t$ obtained from Equation \ref{GAT}. $f_3(\cdot)$ is a linear layer used to project scalar inputs into vector representations. $Attn(\cdot, \cdot, \cdot)$ represents multi-head attention, where the query is $\bm{\bar{z}}_i$, the key is  $\bm{e_i^q}|_{i=1}^{t-1}$, and the value is $\bm{u_i^q}|_{i=1}^{t-1}$. $\bm{s}_i \in \mathbb{R}^d$ is the learning state of the student at step $i$.

With the learning state, we estimate the student's mastery of $\mathcal{C}_i$: 
\begin{equation}
\label{mas_lever}
\begin{aligned}
    k \sim f_m(\bm{s}_i),  \\
    \bm{e}^m_i = \bm{E}^m[k],
\end{aligned}
\end{equation}
where $f_m$ is a function that maps the learning state $\bm{s}_i$ to a probability distribution over discrete mastery levels. We implement it using an MLP followed by a softmax layer. The symbol $\sim$ denotes sampling based on the predicted probabilities. The variable $k$ is an integer ranging from $0$ to $l-1$, representing the selected mastery level. It is used as an index into the mastery level embedding matrix $\bm{E}^m \in \mathbb{R}^{l \times d}$ to retrieve the embedding vector corresponding to the selected level.

Finally, we predict the correctness of the student responding question $q_t$ by:
\begin{equation}
\begin{aligned}
    & h_i=(s_i+e_i^m) \oplus e_i^q\\
    & prob_i=softmax(MLP(h_i)),    
\end{aligned}
\end{equation}
where $p_i$ denotes the predicted probability that the student can correctly answer the current question $q_i$. We set the predicted response to correct if $p_i \ge 0.5$, and incorrect otherwise, that is:
\begin{equation}
    \hat{r}_i = 
    \left\{  
     \begin{array}{lr}  
     1, \text{if }p_i \ge 0.5, &  \\  
     0, \text{otherwise,} &    
     \end{array}  
\right.  
\end{equation}
where $\hat{r}_i$ is the predicted correctness of response.

\paragraph{\textbf{Optimization}} To enable the simulation module to acquire the domain knowledge and reasoning capability of the LLM, we leverage the information in mastery and credit score in the distilled dataset. Specifically, for an arbitrary record at step $t$ in the distilled dataset $D_r$ or pseudo QA (PQA) records, we leverage the QA history $H_t^u$ and the concepts covered by question $q_t$ to compute the learning state $\bm{s}_t$ via Eq.~\ref{DR2}, \ref{learning state} first, and feed $\bm{s}_t$ to a MLP compute the scalar of mastery by:
\begin{equation}
    \hat{m}_i = MLP(\bm{s_i}),
\end{equation}
where $\hat{m}_i$ is the predict mastery in scalar.
Then we encourage the simulation module to acquire the domain knowledge and reasoning capability of the LLM by
mean squared error (MSE) loss between the mastery inferred by LLM and the predicted scalar mastery:
\begin{equation}
\label{equ:loss state1}
    L_b = \sum_i \text{MSE}(m_i, \hat{m}_i).
\end{equation}

With the inspiration of \cite{zhou2025disentangled, long2021tracing}, to encourage the simulation module to make the correct prediction, we constrain the parameters by 
\begin{equation}
\label{equ:loss state2}
    L_s = \beta \cdot L_c + L_p.
\end{equation} 
Here,
\begin{equation}
    L_p = BCE([p_i, 1-p_i], [r_i, 1-r_i]),
\end{equation}
where BCE is Binary Cross-Entropy (BCE) loss.
\begin{equation}
L_c = -\sum_i \mathbb{I}(r_i=\hat{r}_i) \cdot \log{f_m(s_i)[k]},
\end{equation}
where $\mathbb{I}(r_i = \hat{r}_i)$ is an indicator function that equals 1 if $r_i$ equals $\hat{r}_i$, and 0 otherwise. The term $\log{f_m(s_i)[k]}$ represents the probability of the mastery classifier sampling the mastery level $k$.

To maintain training stability, we adopt a two-stage training approach. In the first stage, the model is optimized using Equation \ref{equ:loss state1} with the goal of distilling the knowledge and capabilities of the LLM. In the second stage, the model is optimized using Equation \ref{equ:loss state2} to enable it to generate responses based on the distilled knowledge and capabilities.

\section{Experimental Evaluation}

\renewcommand\arraystretch{1.5}
\begin{table*}[t]
    \centering

    \caption{ACC and AUC of different QA simulators on 4 datasets. \textbf{Bold} indicates the best performance among all QA simulators. \underline{Underline} indicates the second-best performance. * indicates p-value < 0.01 in the significance test. - indicates the AUC of Agent4Edu is unavailable because it directly prompts an LLM to output ``yes'' or ``no''.
    }
    \label{tab:main result}
    \scriptsize
    \begin{lrbox}{\tablebox}

    \setlength{\tabcolsep}{2.7mm} {
    \begin{tabular}{c| c|c c|c c|c c|c c}
     \hline

        \multicolumn{2}{c|}{\multirow{2}*{\textbf{ }}} &
        \multicolumn{2}{c|}{\textbf{Junyi}} &
        \multicolumn{2}{c|}{\textbf{Assist09}} &
        \multicolumn{2}{c|}{\textbf{Assist12}} &
        \multicolumn{2}{c}{\textbf{Algebra}}
        \\
            \cline{3-10}
            
        \multicolumn{2}{c|}{\multirow{2}*{\textbf{ }}} & 
        ACC & AUC & ACC & AUC & ACC & AUC & ACC & AUC \\
            \hline
            
        \multirow{7}*{
        \textbf{LLM-free}
        } & DKT &
        0.6365 & 0.6485 & 
        0.4728 & 0.6278 & 
        0.4806 & 0.5940 & 
        0.4009 & 0.6218 \\
            \cline{2-10}

        & AKT &
        0.6494 & 0.7467 & 
        0.5693 & 0.6926 & 
        0.4572 & 0.6070 & 
        0.5782 & 0.6805 \\
            \cline{2-10}

        & ATKT &
        0.5104 & 0.6693 & 
        0.6040 & 0.5646 & 
        0.4654 & 0.6055 & 
        0.4937 & 0.6730 \\
            \cline{2-10}

        & SAKT &
        0.7664 & 0.7122 & 
        0.4303 & 0.6442 & 
        0.6930 & 0.6522 & 
        0.5550 & 0.6765 \\
            \cline{2-10}

        & Deep-IRT &
        0.6692 & 0.6524 & 
        0.4493 & 0.6081 & 
        0.5141 & 0.6061 & 
        0.5089 & 0.6584 \\
            \cline{2-10}

        & DisKT &
        0.5155 & 0.7247 & 
        0.6216 & 0.7241 & 
        0.6009 & 0.6673 & 
        0.5043 & 0.6752 \\
            \cline{2-10}

        & DSim &
        0.7789 & 0.7486 & 
        0.6702 & 0.7149 & 
        0.7190 & 0.6570 & 
        0.7880 & 0.6552 \\
            \cline{1-10}

        \multirow{3}*{
        \textbf{LLM-based}
        } & LLM-KT &
        0.7671 & 0.4898 & 
        0.6027 & 0.4767 & 
        0.7150 & 0.5130 & 
        0.7836 & 0.5187 \\
            \cline{2-10}

        & Agent4Edu &
        0.7513 & - & 
        0.6543 & - & 
        0.6877 & - & 
        0.7617 & - \\
            \cline{2-10}

        & SinKT &
        \underline{0.7952} & \underline{0.7852} & 
        \underline{0.6849} & \underline{0.7436} & 
        \underline{0.7264} & \underline{0.7112} & 
        \underline{0.8001} & \underline{0.7725} \\
            \cline{1-10}

        \textbf{ours} & LDsim &
        \textbf{0.8179*} & \textbf{0.8668*} & 
        \textbf{0.8260*} & \textbf{0.8991*} & 
        \textbf{0.7887*} & \textbf{0.8340*} & 
        \textbf{0.8457*} & \textbf{0.8545*} \\
            \cline{1-10}

	\end{tabular}  }
    \end{lrbox}
    \scalebox{1.2}{\usebox{\tablebox}}
\end{table*}

In this section, we conduct experiment to evaluate the performance of LDSim. Our code
will be released in \url{https://anonymous.4open.science/r/LDSim-05A9}.

\subsection{Dataset}
We evaluate the performance of our method with the following datasets:

\begin{itemize} [leftmargin=15pt]
    \item \textbf{Junyi} \footnote{https://www.kaggle.com/datasets/junyiacademy/learning-activity-public-dataset-by-junyi-academy} is a dataset of QA records collected by the Junyi Academy online education platform \footnote{https://www.junyiacademy.org} during the 2018-2019 academic year.

    \item \textbf{Assist09} \footnote{\url{https://sites.google.com/site/assistmentsdata/home/2009-2010-assistment-data}} is a dataset of student QA records collected by the ASSISTments online education platform \cite{feng2009addressing} during the 2009-2010 academic year.

    \item \textbf{Assist12} \footnote{\url{https://sites.google.com/site/assistmentsdata/datasets/2012-13-school-data-with-affect}} is a dataset of student QA records collected by the ASSISTments online education platform \cite{feng2009addressing} during the 2012-2013 academic year.

    \item \textbf{Algebra} \footnote{https://www.kdd.org/kdd-cup/view/kdd-cup-2010-student-performance-evaluation/Data} is a challenge dataset provided by KDD CUP 2010 \footnote{https://kdd.org/kdd-cup/view/kdd-cup-2010-student-performance-evaluation/Intro}, which contains a large number of QA records from students.
\end{itemize}

\renewcommand\arraystretch{1.3}
\begin{table}[t]
    \centering
    \caption{The statistics of datasets.}
    \label{tab:dataset}

    \scriptsize

    \begin{lrbox}{\tablebox}
    \begin{tabular}{c|c|c|c|c}
     \hline
        \textbf{Dataset} & Junyi & Assist09 & Assist12 & Algebra   \\
            \hline

        \textbf{QA record number} & 2525877 & 258461 & 825548 & 989189 \\
            \hline

        \textbf{student number} & 9879 & 2186 & 5183 & 5047  \\
            \hline

        \textbf{question number} & 9811 & 12625 & 15575 & 13994 \\
            \hline

        \textbf{concept number} & 574 & 139 & 103 & 692 \\
            \hline

        \textbf{max concepts per question} & 1 & 6 & 1 & 6 \\
            \hline

        \textbf{Positive Label Rate} & 73.74\% & 63.34\% & 70.45\% & 80.68\% \\
            \hline
 
	\end{tabular}
    \end{lrbox}

    \scalebox{1.1}{\usebox{\tablebox}}
\end{table}

For these four datasets, we uniformly set the maximum QA history length per student to 200. Since the Assist09 dataset is relatively small, we only remove records of students with fewer than 50 QA records, whereas for the other datasets, we remove records of students with fewer than 100 QA records. After these operations, the basic information of our datasets is shown in the Table \ref{tab:dataset}. We split each dataset into training, validation, and test sets in an 8:1:1 ratio for our experiments.

\subsection{Baselines}
To validate the effectiveness of our method, we selected both LLM-free methods and LLM-based methods as our baseline. For the LLM-free methods, we select:
\begin{itemize}[leftmargin=15pt]
    \item \textbf{DKT} \cite{piech2015deep} is a classic model that use RNNs to model students' QA history and to predict their responses.
    
    \item \textbf{AKT} \cite{ghosh2020context}, \textbf{ATKT} \cite{guo2021enhancing}, \textbf{SAKT} \cite{pandey2019self} are classic models that use attention mechanisms to model students' QA history and to predict their responses.
    
    \item \textbf{Deep-IRT} \cite{yeung2019deep} is a model that combines Item Response Theory (IRT) with deep learning to model students' QA history and to predict their responses.
    
    \item \textbf{DisKT} \cite{zhou2025disentangled} is a recently proposed model that focuses on alleviating cognitive bias present in previous models.
    
    \item \textbf{DSim} \cite{long2025simulating} is a recently proposed model that focuses on addressing the bias accumulation problem found in previous simulation task.
\end{itemize}
Then, for LLM-based methods, we have:
\begin{itemize}
    \item \textbf{LLM-KT} \cite{wang2025llm} is fine-tuning based method that leverages the knowledge and capabilities of LLMs to make prediction. 
    
    \item \textbf{Agent4Edu} \cite{gao2025agent4edu} is a prompt engineering based method that make prediction by prompting the LLM agent..
    
    \item \textbf{SINKT} \cite{fu2024sinkt} is a method based on leveraging LLMs' domain knowledge to encode questions and concepts in order to enhance the prediction.
\end{itemize}

\subsection{Implementation Details}
In our experiments, we set $n=30$ in Eq. \ref{prob_def}, meaning that for each student, QA records prior to the last 30 steps are regarded as the student's QA history. 

We apply GLM-4-Flash \cite{du2021glm} as our foundation LLM. The head of the attention in Eq.\ref{learning state} is 1.
Regarding hyperparameters, we set the embedding dimension $d$
 for questions and concepts to 128, and $\beta$ in Equation \ref{equ:loss state2} to 40. We optimize the model using Adam \cite{kingma2014adam} with a learning rate of 0.001. For baselines that only require prompting an LLM via API calls, we uniformly adopt GLM-4-Flash as the base LLM model. Since our datasets contain only textual information of concepts, for baselines needing additional textual inputs, we exclude those parts. For other settings of baselines, we follow the original papers and official code settings.

\subsection{Overall Performance}
We compare the prediction accuracy (ACC) and the Area Under Curve (AUC) of LDSim with the baselines.
The results are shown in the Table \ref{tab:main result}. As  Agent4Edu directly generates the predicted responses via prompting an LLM, and therefore its AUC cannot be computed. From the Table \ref{tab:main result}, we can observe that:

(1) LDSim outperforms all the baselines across all cases in the simulation task, achieving a $2\%-20\%$ improvement over the current SOTA methods in different cases, which strongly demonstrates the superiority of our approach.

(2) Compared with LLM-free simulators, LLM-based simulators achieve overall higher performance. This is because LLM-free simulators cannot effectively leverage the semantic information of concepts in the data and cannot benefit from the reasoning capabilities of LLMs. In contrast, LLM-based simulators either possess rich world knowledge or strong reasoning capabilities, which can effectively enhance the ability to predict student responses.

(3) Although both LLM-based methods and our LDSim leverage the domain knowledge and capabilities of LLMs, the baselines still underperform. We hypothesize that this is because our model distills and refines only the domain knowledge and reasoning abilities of the LLM that are directly relevant to the simulation task. In contrast, the baselines rely on the LLM's more general knowledge and capabilities, which may introduce noise from irrelevant information.

\subsection{Deployment Cost}
To investigate the deployment cost of LDSim,  we compare the time and computational resource costs of LDSim with the LLM-based method during deployment. Specifically, we select a student with 200 QA records and task each simulator with predicting the last 30 responses to simulate the student QA behavior. We then record the time taken and GPU memory usage for completing this task. The results are shown in the Table \ref{tab:cost}. Since Agent4Edu calls the LLM API to generate responses, its actual memory usage is unavailable. From the Table \ref{tab:cost}, we can observe that: (1) LLM-based simulators exhibit extremely slow response times, requiring from 10 seconds to nearly 50 minutes to simulate a single student, which is an impractical cost in ERS training scenarios. In contrast, our LDSim completes the simulation of a student's 30-step responses in just 0.73s. (2) The LLM fine-tuning based method demands substantial computational resources for deployment,
 whereas SinKT and LDSim have much lower GPU requirements. Although LDSim consumes more GPU memory than SinKT, its ultra-fast response time makes this memory cost acceptable.

\renewcommand\arraystretch{1.7}
\begin{table}[t]
    \centering
    \caption{The time and computational resource costs}
    \label{tab:cost}

    \scriptsize

    \begin{lrbox}{\tablebox}
    \begin{tabular}{c|c|c|c}
     \hline
         \multicolumn{2}{c|}{} & Time Cost(s) & Memory Cost(MB)   \\
            \hline

        \multirow{3}*{
        \textbf{LLM-based}
        } & LLM-KT & 170.24 & 2338.01 \\
            \cline{2-4}

        & Agent4Rdu & 3117.62 & - \\
            \cline{2-4}

        & SinKT & 12.30 & 66.43 \\
            \hline

        \textbf{ours} & LDSim & 0.73 & 170.28 \\
            \hline
 
	\end{tabular}
    \end{lrbox}

    \scalebox{1.1}{\usebox{\tablebox}}
\end{table}

\subsection{Ablation Study}
\begin{figure}
    \centering
    \includegraphics[width=1.0\linewidth]{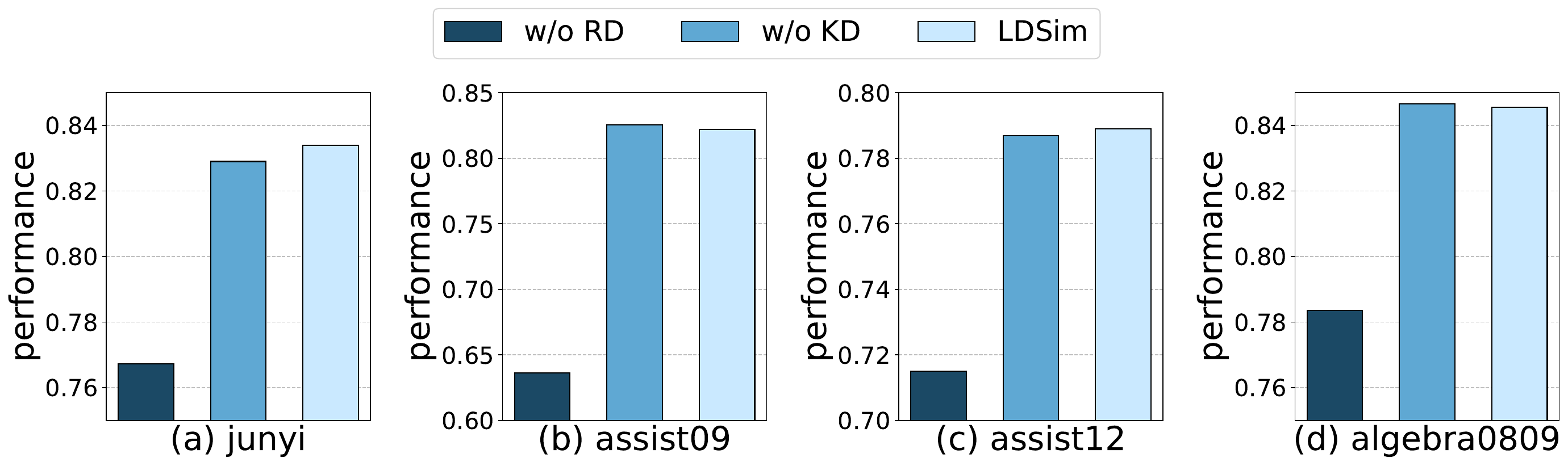}
    \caption{Ablation Study}
    \label{fig:ablation}
\end{figure}

To further investigate the contribution of each module in LDSim, we conducted an ablation study by removing the world knowledge distillation module (w/o KD) and the reasoning capability distillation module (w/o RD). For w/o KD, we replaced the concept relationship graph $\mathcal{G}$ generated by the LLM with a fully connected graph. For w/o RD, we skipped the LLM-generated mastery information and omitted the first-stage training of the mastery estimator. 
The results are shown in the Figure \ref{fig:ablation}.

From the Figure \ref{fig:ablation} we can observe: (1) In most cases, removing any of the two components results in a decrease in LDSim's overall performance, indicating that each component contributes positively to the model's effectiveness. (2) The removal of the reasoning capability distillation module causes the largest performance drop, suggesting that the LLM's reasoning ability provides the most significant performance gain for LDSim. (3) The world knowledge distillation module contributes less to the performance gains of LDSim, possibly because the mastery generated by the LLM already contain some world knowledge, which is partially distilled during the reasoning capability distillation process.

\subsection{Case Study}
\begin{figure}
    \centering
    \includegraphics[width=0.95\linewidth]{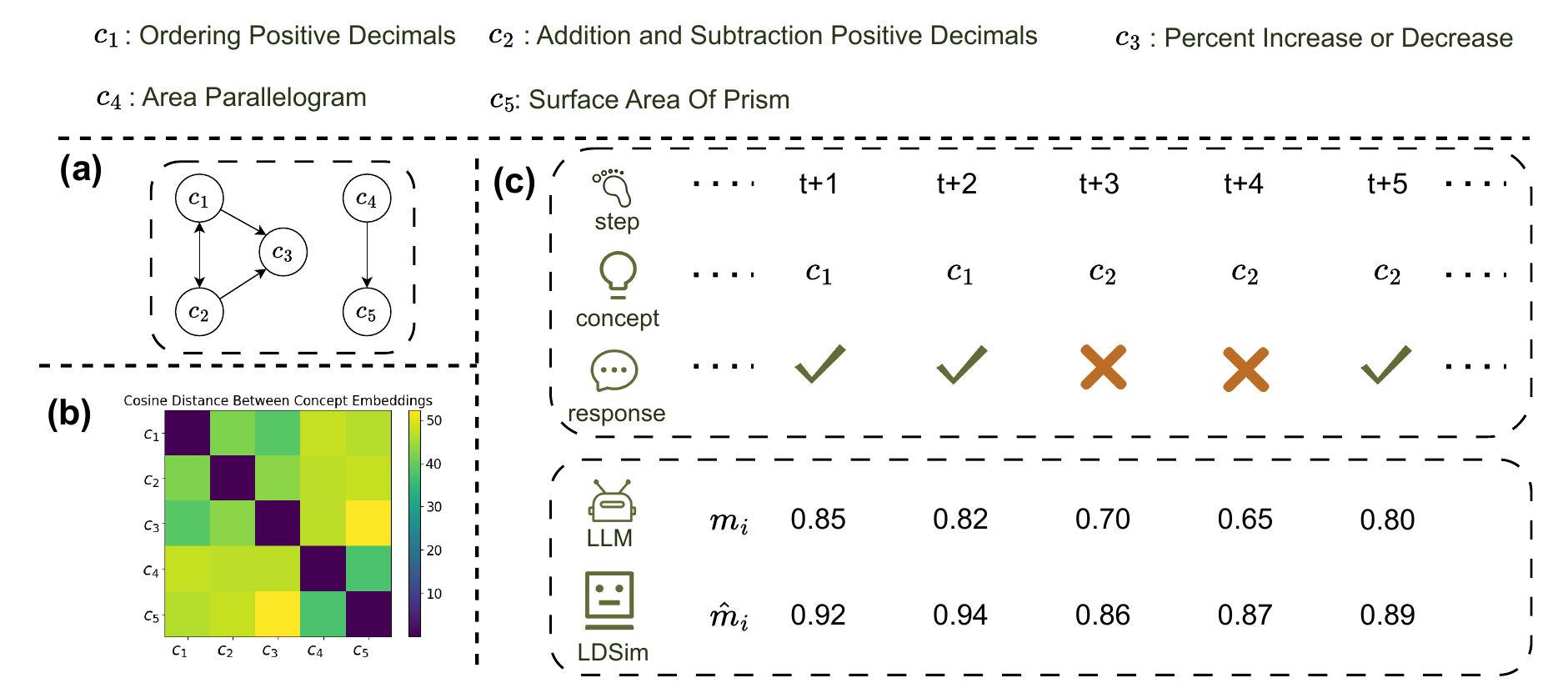}
    \caption{Case study}
    \label{fig:case study}
\end{figure}

To further investigate whether we have effectively distilled the LLM, and whether LDSim effectively leverages them, we conducted a case study. Specifically, we take Assist12 as an example to illustrate how LDSim distills and leverages LLMs

First, we examined the knowledge distillation process. Figure \ref{fig:case study}(a) shows the relationships among five concepts in the concept graph $\mathcal{G}_c$ generated by the KD module. We can observe that the module not only successfully identifies the correlations between concepts (distinguishing between algebra and geometry knowledge) but also accurately captures the prerequisite relationships among concepts (indicating that a student should master “Area Parallelogram” before “Surface Area of Prism”). This demonstrates that we have effectively distilled the knowledge of LLMs.
Next, we verify whether this distilled knowledge is effectively utilized by the simulation module. Figure \ref{fig:case study}(b) presents the cosine similarities among the semantic embeddings of these five concepts in a trained LDSim. We observe that, after being encoded by the GAT in Eq. \ref{GAT}, the semantic embeddings of correlated concepts are more similar than those of uncorrelated ones. For instance, the similarity between \textit{Ordering Positive Decimals}($c_1$) and \textit{Percent Increase or Decrease}($c_3$) is higher than the similarity between \textit{Percent Increase or Decrease}($c_3$) and \textit{Area Parallelogram}($c_5$). 
This indicates that the distilled knowledge has been effectively encoded into the simulation module.

\renewcommand\arraystretch{1.5}
\begin{table*}[t]
    \centering

    \caption{ACC and AUC of QA simulators under the single-step simulation setting on the four datasets. \textbf{Bold} indicates the best performance among all QA simulators. \underline{Underline} indicates the second-best performance. * indicates p-value < 0.01 in the significance test.}
    \label{tab:single step}
    \scriptsize
    \begin{lrbox}{\tablebox}

    \setlength{\tabcolsep}{2.7mm} {
    \begin{tabular}{c| c|c c|c c|c c|c c}
     \hline

        \multicolumn{2}{c|}{\multirow{2}*{\textbf{ }}} &
        \multicolumn{2}{c|}{\textbf{Junyi}} &
        \multicolumn{2}{c|}{\textbf{Assist09}} &
        \multicolumn{2}{c|}{\textbf{Assist12}} &
        \multicolumn{2}{c}{\textbf{Algebra}}
        \\
            \cline{3-10}
            
        \multicolumn{2}{c|}{\multirow{2}*{\textbf{ }}} & 
        ACC & AUC & ACC & AUC & ACC & AUC & ACC & AUC \\
            \hline
            
        \multirow{7}*{
        \textbf{LLM-free}
        } & DKT &
        0.7516 & 0.7162 & 
        0.6967 & 0.7427 & 
        0.7302 & 0.7023 & 
        0.8258 & 0.7976 \\
            \cline{2-10}

        & AKT &
        0.8010 & 0.7954 & 
        0.7170 & 0.7650 & 
        \underline{0.7507} & \underline{0.7514} & 
        \underline{0.8293} & \underline{0.7992} \\
            \cline{2-10}

        & ATKT &
        0.7365 & 0.6886 & 
        0.6473 & 0.5630 & 
        0.7192 & 0.6663 & 
        0.8240 & 0.7912 \\
            \cline{2-10}

        & SAKT &
        0.7360 & 0.7104 & 
        0.6868 & 0.7289 & 
        0.7268 & 0.6877 & 
        0.8233 & 0.7860 \\
            \cline{2-10}

        & Deep-IRT &
        0.7494 & 0.7211 & 
        0.6956 & 0.7324 & 
        0.7306 & 0.6938 & 
        0.8276 & 0.7950 \\
            \cline{2-10}

        & DisKT &
        0.7921 & 0.7911 & 
        0.7183 & 0.7690 & 
        0.7294 & 0.7182 & 
        0.8184 & 0.7791 \\
            \cline{2-10}

        & DSim &
        0.7789 & 0.7486 & 
        0.6702 & 0.7149 & 
        0.7190 & 0.6570 & 
        0.7880 & 0.6552 \\
            \cline{1-10}

        \multirow{3}*{
        \textbf{LLM-based}
        } & LLM-KT &
        - & - & 
        0.6174 & 0.5044 & 
        0.7058 & 0.5052 & 
        0.8055 & 0.5421 \\
            \cline{2-10}

        & Agent4Edu &
        0.7592 & - & 
        0.6553 & - & 
        0.7088 & - & 
        0.7849 & - \\
            \cline{2-10}

        & SinKT &
        \underline{0.8022} & \underline{0.7997} & 
        \underline{0.7287} & \underline{0.7811} & 
        0.7416 & 0.7418 & 
        0.8251 & 0.7852 \\
            \cline{1-10}

        \textbf{ours} & LDsim &
        \textbf{0.8217*} & \textbf{0.8730*} & 
        \textbf{0.8229*} & \textbf{0.8966*} & 
        \textbf{0.8022*} & \textbf{0.8528*} & 
        \textbf{0.8646*} & \textbf{0.8712*} \\
            \cline{1-10}

	\end{tabular}  }
    \end{lrbox}
    \scalebox{1.2}{\usebox{\tablebox}}
\end{table*}

\subsection{Single-Step Simulation}
In certain educational recommendation scenarios, the ERS can immediately receive a student's response after recommending a question, allowing it to adjust the next recommendation in real time. To train the ERS under such scenarios, we evaluated the simulators' ability to predict students' single-step responses. In this setting, given all QA records prior to the current step $\{(q_t, \mathcal{C}_t, r_t)|_{t=1}^{i-1}\}$, the QA simulator needs to predict the student's response $r_i$ at the current step. We conducted experiments for all QA simulators on the four datasets and recorded the ACC and AUC metrics. The results are shown in the Table \ref{tab:single step}.

From the Table, we can observe that (1) our LDSim still outperforms all baselines in the single-step simulation setting, achieving $2\%$ to $15\%$ improvement over the current state-of-the-art methods. This demonstrates that our QA simulator can effectively train ERS in more general educational scenarios. (2) Compared to the multi-step simulation setting, QA simulators generally perform better in the single-step setting. This is because, in multi-step simulations, the simulator does not have access to the true responses of the last few QA records, so any prediction errors accumulate over subsequent steps. In the single-step setting, such cumulative errors do not occur.

\bibliographystyle{ACM-Reference-Format}
\bibliography{sample-base}

\end{document}